\documentclass[letterpaper, 10 pt, conference]{ieeeconf}  

\IEEEoverridecommandlockouts                              
\usepackage{amssymb}
\usepackage{graphicx,amsmath}
\usepackage{epstopdf}
\usepackage{lineno}
\usepackage{caption,url}
\usepackage{float}

\usepackage{subcaption}
\usepackage{gensymb}
\usepackage[colorlinks]{hyperref}
\hypersetup{
    colorlinks,
    linkcolor=black,
    citecolor=black,
    filecolor=black,
    urlcolor=black,
}

\title{\LARGE \bf
Imitation Learning for High Precision Peg-in-Hole Tasks
}

\author{Sagar Gubbi\textsuperscript{*}, Shishir Kolathaya\textsuperscript{*}, Bharadwaj Amrutur 
\thanks{\noindent * S. Gubbi and S. Kolathaya have contributed equally in this manuscript.}
\thanks{This work is supported by Yaskawa India Pvt. Ltd.}
\thanks{S. Gubbi is a PhD student, S. Kolathaya and B. Amrutur are with the faculty of the Robert Bosch Center for Cyber-Physical Systems, Indian Institute of Science, Bengaluru, India. {\tt\small email: \{sagar,shishirk,amrutur\} at iisc.ac.in}}%
}%

\begin{document}
\maketitle
\thispagestyle{empty}
\pagestyle{empty}



\begin{abstract}
Industrial robot manipulators are not able to match the precision and speed with which humans are able to execute contact rich tasks even to this day. 
Therefore, as a means overcome this gap, we demonstrate generative methods for imitating a peg-in-hole insertion task in a $6$-DOF robot manipulator.
In particular, generative adversarial imitation learning (GAIL) is used to successfully achieve this task with a 
$6~\mu$m peg-hole clearance on the Yaskawa GP8 industrial robot. Experimental results show that the policy successfully learns within $20$ episodes from a handful of human expert demonstrations on the robot (i.e., $< 10$ tele-operated robot demonstrations). The insertion time improves 
from $ > 20$ seconds (which also includes failed insertions) to $ < 15$ seconds, 
 thereby validating the effectiveness of this approach.

\end{abstract}




\section{Introduction}

\begin{figure}\centering
  \includegraphics[height=0.25\textwidth]{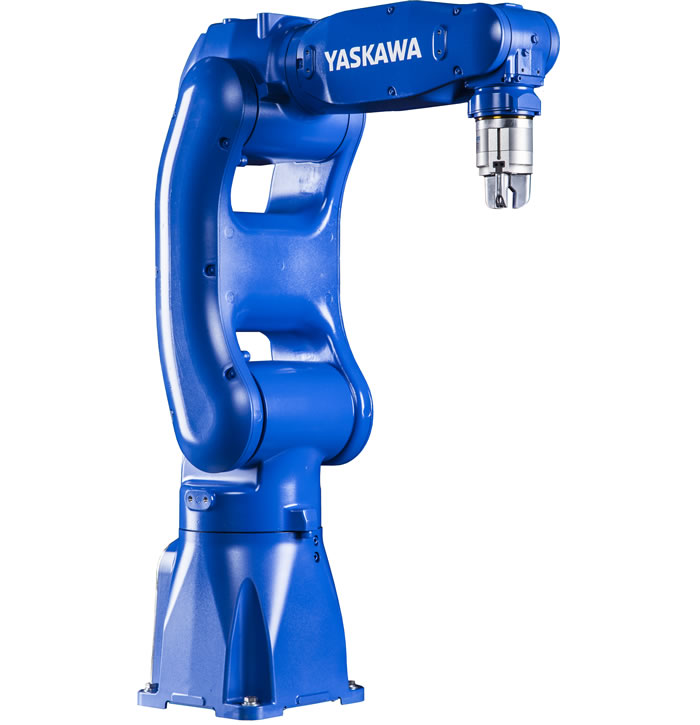}
  \includegraphics[height=0.25\textwidth]{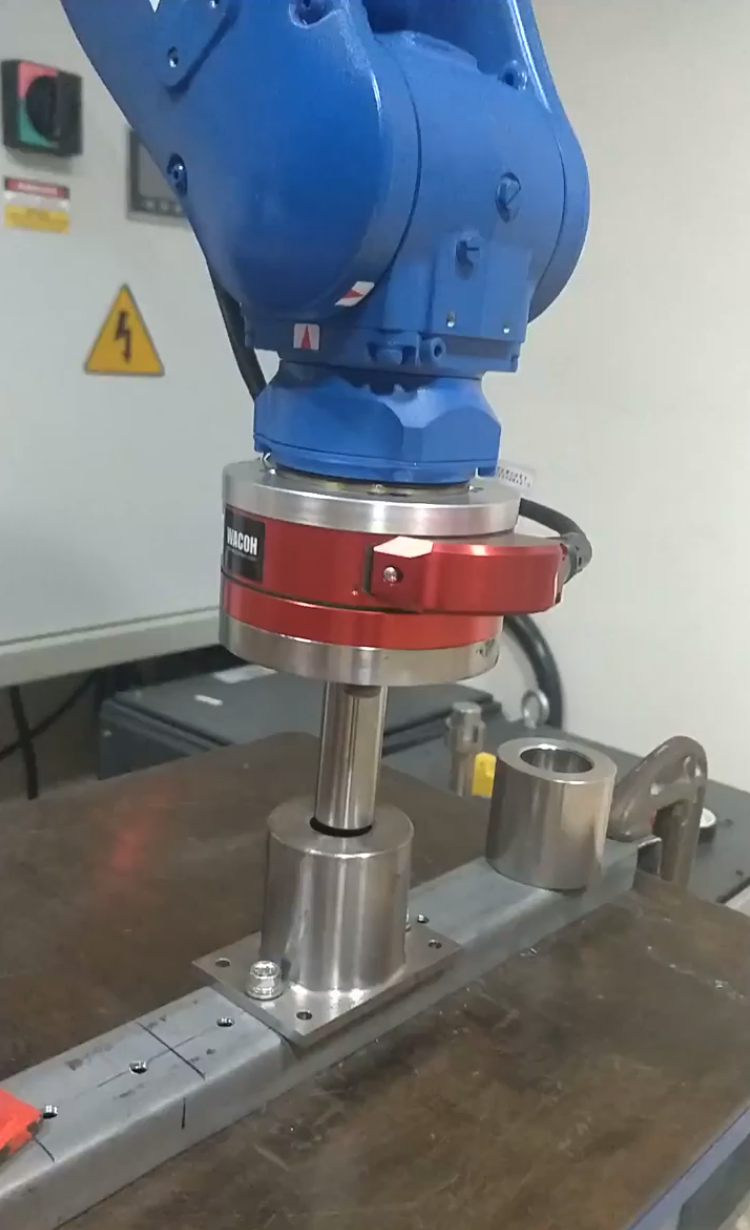}
  \caption{The Yaskawa GP8 robot is shown on the left, and the peg being inserted by this robot into a hole with a gap of 6~$\mu m$ is shown on the right. }
  \label{fig:sixandtenmicrons}
\end{figure}

High precision assembly tasks refer to the class of tasks where the position accuracy required far exceeds that of the robot manipulators. For example, the Yaskawa GP8 robot (shown in Fig. \ref{fig:sixandtenmicrons}) has accuracy levels of $20-100~\mu$m, whereas some of the peg-in-hole insertion tasks require accuracy levels of $\leq 10~\mu$m. More importantly, these types of insertion tasks are contact-rich (as shown by Fig. \ref{fig:sixandtenmicrons}), thereby requiring limits on the maximum allowable forces exerted by the peg. Therefore, executing these types of tasks remain a challenge for traditional control schemes even to this day.

Classical methods used for the peg-in-hole task involve the application of a transimpedance controller rather than a direct position or velocity control of the joint angles \cite[Chapter 8.8]{ghosal2006robotics}. A transimpedance controller realizes active compliant behavior of the end-effector, thereby enabling peg-in-hole insertions despite the lack of positioning accuracy of the manipulators. It is important to note that the forces normal to the surface must be controlled to prevent any form of damage to the peg or the hole. However, even if the stiffness and damping parameters of the impedance controller are well tuned, the insertion times are high ($>20$ seconds) for tight clearances ($\leq10~\mu$m). This is in contrast to insertion times taken by human experts, which are $<5$ seconds. Hence, the question that we would like to answer is, can we mimic this human expert policy in the robot via one of the well known imitation learning methods available in literature?

It is worth noting that modeling the contact is difficult, but the policy employed by human experts to insert the peg is seemingly simple. 
In imitation learning, a policy network is trained to copy the expert and take the same actions that the expert would have taken given the same observations \cite{zhang2018deep, ho2016generative}. It is a robust skill acquisition method that has been successfully applied to autonomous driving \cite{bojarski2016end} and complex manipulation tasks \cite{yu2018one}. Therefore, the goal of the paper is to mimic an expert policy via imitation learning for the peg-in-hole task.

Imitation learning requires expert data. In some domains, such as autonomous driving, the control interface makes this straight forward to procure. The observations from the camera and other sensors can be recorded along with the driver's actions on the steering wheel, accelerator, and brake pedals. Expert data can also be collected for complex manipulation tasks by teleoperating a robot with a VR controller \cite{zhang2018deep}. For high precision assembly tasks, it is much harder to collect expert data. Although humans can reliably place a peg in a hole, it is difficult to capture the forces applied and the feedback forces received. Hence, we chose to gather useful expert data for the peg-in-hole task by teleoperation. Since the amount of expert data is limited in this context, the question we address in this work is: Can imitation learning succeed in copying an expert with a small set of demonstrations?

\begin{figure}
	\centering
	\includegraphics[width=0.5\columnwidth]{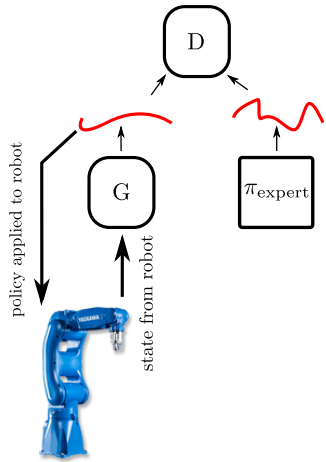}
	\caption{Figure showing the GAIL framework used for the peg-in-hole task. The generator is denoted by G, and the discriminator is denoted by D, and the expert policy is denoted by $\pi_{\rm{expert}}$.}
	\label{fig:gangp8}
\end{figure}

To address this issue, we first build a teleoperation system that uses a space mouse (see Fig. \ref{fig:spacemouse_vj}) to control a Yaskawa GP8 robot. The feedback forces received by the MotoFit force sensor are plotted on a display, and the expert can exert forces on the peg using the space mouse. The feedback forces and the applied forces are recorded as the expert performs the peg-in-hole task. The recorded data is used to train a neural network that takes the force feedback and the current position and produces the force to be applied. We specifically use generative adversarial imitation learning (GAIL) \cite{ho2016generative} for learning the policy, which are known to be sample efficient. Pictorial representation of GAIL is provided in Fig. \ref{fig:gangp8}.

Our contributions are:
\begin{itemize}
    \item We show that a neural network can be trained via generative methods to copy the actions taken by a human expert.
    \item We find that only a handful of expert trajectories (less than ten) are sufficient for the peg-in-hole task to achieve a high success rate.
\end{itemize}

The remainder of this paper is structured as follows. In Section \ref{sec:relatedwork}, we discuss related work. Following this in Section \ref{sec:arch}, we describe the proposed architecture. Subsequently, we present experimental results in Section \ref{sec:results} and conclude this paper with a short discussion in Section \ref{sec:conc}.

\section{Related Work}
\label{sec:relatedwork}

Deep Reinforcement learning (D-RL) has been used to learn controllers for a variety of tasks ranging from walking robots \cite{Dholakiya2019, Bhattacharya2019, Kolathaya2019} to manipulating objects with an arm \cite{lillicrap2015continuous, gu2017deep, levine2013guided, popov2017data, pinto2016supersizing}. Hence reinforcement learning, indeed, offers a way to 
realize peg-in-hole tasks via random explorations, 
thereby eliminating the need to hand craft an effective control/policy without using any form of expert data. 
However, the remarkable success observed in D-RL cannot be translated directly to realize high precision assembly tasks.
One of the factors being the hard limit on the number of allowable training iterations in hardware. As the tasks become more complex, the training becomes more expensive.
Sample efficiency of the learning algorithm is critical to directly deploy on real robots \cite{haarnoja2018soft}. 
For example, in \cite{inoue2017deep}, the task of inserting the peg inside the hole 
defines a reward function and uses Q learning to arrive at a policy that, given the measured feedback forces, outputs the forces to be applied. The minimum number of episodes required for learning a stable policy for a peg-hole clearance of $10~\mu$m was $100$. Therefore, with a view toward reducing the number of training episodes, the goal of this paper is to use imitation learning to copy the policy of human experts. 
We find that imitation is significantly more sample efficient; uses $<20$ episodes to fully learn the policy even for pegs with tighter clearances (i.e., $6~\mu$m).



\cite{tang2016autonomous} proposed to use an analytical model of the contact to develop better policies for the peg-in-hole insertion tasks.
The reactionary forces when a peg is pushed against the hole at an arbitrary angle are recorded, and the model is used to solve for the actual orientation of the peg. Once the orientation is known, the peg is pushed in at the appropriate angle. In our work, we assume the peg is aligned with the hole within $2$ degrees and obtain a policy that can push the peg inside the hole and react appropriately if the peg is stuck mid way through the hole. We do not attempt to model the contact, so our method is model-free.


Visual imitation learning was proposed in \cite{gubbi2019, zhang2018deep, yu2018one, yu2018two} to copy an expert performing complex manipulation tasks using a virtual reality controller. The tasks accomplished include inserting blocks into shape sorting cubes. Although our work is superficially similar, there are several differences. First, we do not use vision because we assume that the peg is close to the hole at the starting position. Second, our objective requires significantly higher precision because the gap between the peg and the hole is only $6~\mu$m. This necessitates the use of impedance control using a force sensor rather than position/velocity control.


\section{Proposed Architecture}
\label{sec:arch}

\begin{figure}
  \includegraphics[width=0.75\linewidth, keepaspectratio]{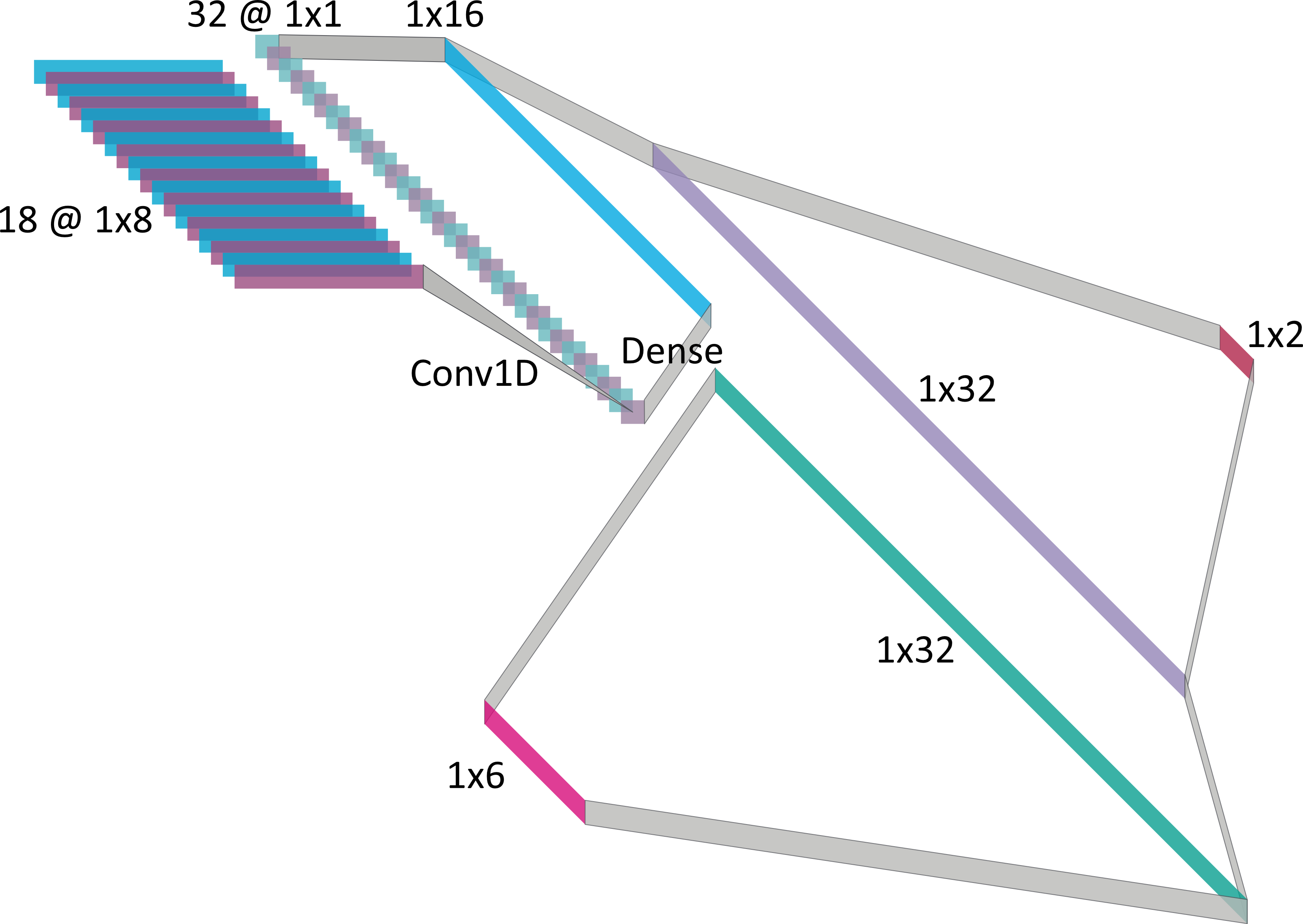}
  \caption{The discriminator network that takes (state, action) pairs as input and classifies if the pair comes from expert data or from the generator (policy) network.}
  \label{fig:disc_nn}
\end{figure}

\begin{figure}
  \includegraphics[width=0.85\linewidth, keepaspectratio]{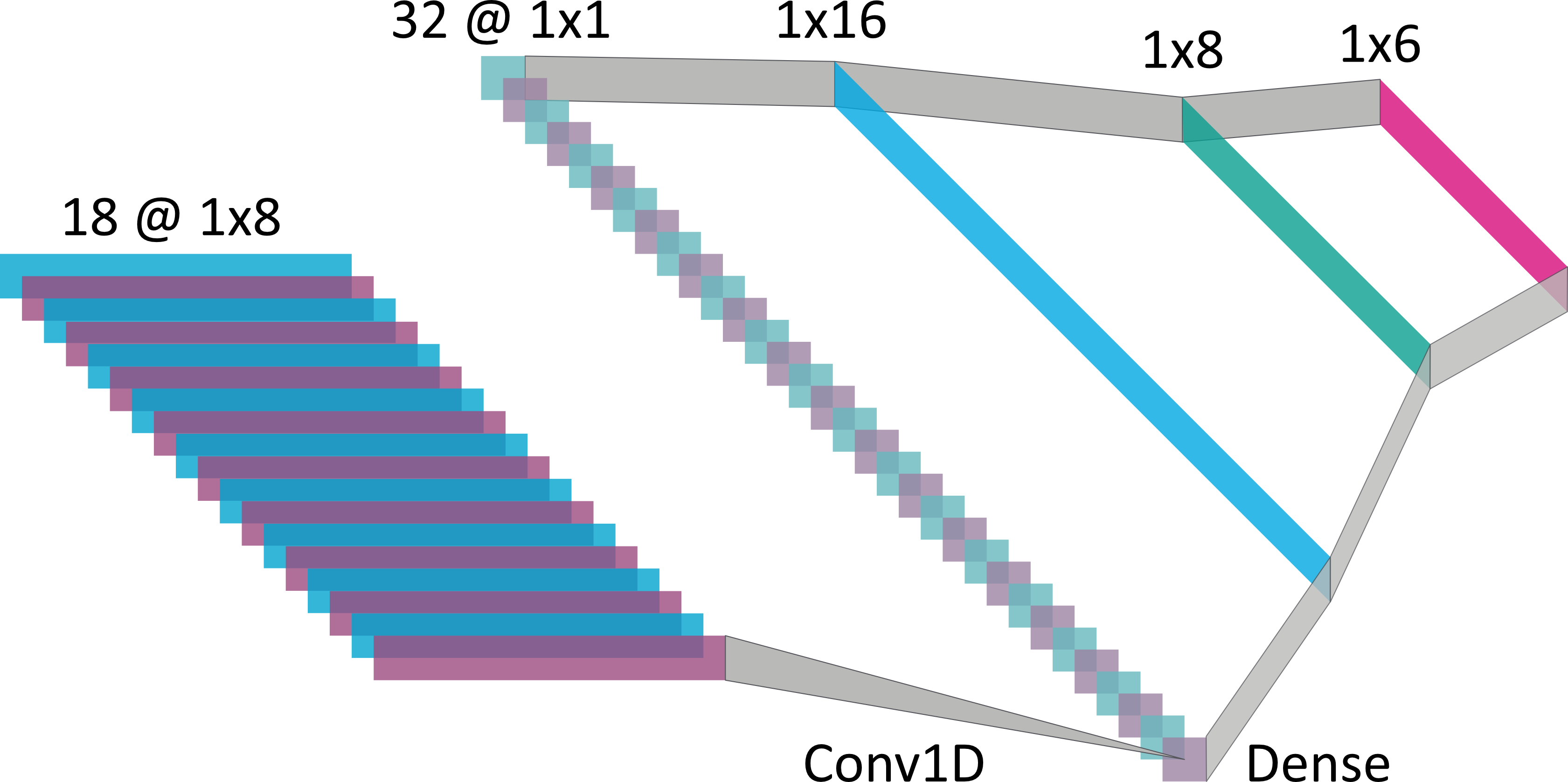}
  \caption{The policy network (generator) that takes the current state (current force feedback and position) and outputs the actuator command (target forces) that are fed as input to the transimpedance controller}
  \label{fig:gen_nn}
\end{figure}

In imitation learning, behavior cloning \cite{zhang2018deep} and generative adversarial imitation learning (GAIL) \cite{ho2016generative} are two approaches to train a policy network. In behavior cloning, once the dataset is collected, it is used to train a policy network through supervised learning, and the robot does no further exploration during learning. 
In GAIL 
the reward function is inferred from expert data, and the inferred reward function is used to train the controller. 
GAIL has the potential to enable the robot to generalize better while using limited training data by training the policy network with reinforcement learning \cite{ho2016generative}. 
This is similar to inverse reinforcement learning where a reward function is extracted from the expert data. However, in GAIL, the reward function is never explicitly recovered from the expert data. Instead, a discriminator network (Fig.~\ref{fig:disc_nn}) is trained to distinguish between expert trajectories and trajectories generated by the generator policy network (see Fig.~\ref{fig:gangp8}). The trained discriminator is then used as a reward function to train the generator network (for example, with PPO \cite{schulman2017proximal}) so as to confuse the discriminator. Thus, training the generator and discriminator proceed alternately in GAIL, and the generator improves over time.
Neural network models for both the generator and discriminator are shown in Figs. \ref{fig:gen_nn}, \ref{fig:disc_nn}.
See \cite{ho2016generative} for more details for the detailed description for the control algorithm for GAIL implemented in the robot. 

After successful training, the generator network becomes our policy network.
The inputs to the policy network are:
\begin{itemize}
\item $x,y,z$ positions of the peg: $p_x$, $p_y$, $p_z$.
    \item Roll, pitch and yaw positions of the peg:
    $p_{r_x}$, $p_{r_y}$, $p_{r_z}$.
\end{itemize}

The output of the policy network determines if the peg should apply a downward force, apply wiggle in every time step. 
Therefore, the action ($A_p$) is a discrete variable, i.e., wiggle\footnote{Here wiggle means random forces being applied along roll, pitch and yaw directions.} with a downward force, wiggle without a downward force, apply a downward force without wiggle, and apply a downward force with wiggle. Hence, there are four actions. We will number the actions from $0$ to $3$ with each action described as follows:
\begin{itemize}
	\item $0$: Downward force with wiggle
	\item $1$: Downward force without wiggle
	\item $2$: Wiggle without downward force
	\item $3$: No wiggle and no downward force
\end{itemize}

The policy network is shown in Fig.~\ref{fig:gen_nn}. The first layer is the LayerNorm normalization layer and is followed by a $1$D convolutional layer along the time axis which accepts the state input from the last few time steps. This is followed by a few densely connected layers to produce the discretized target force outputs.
There are two notable aspects to this network:
\begin{itemize}
    \item At the very beginning, the first layer is a normalization layer with large epsilon. We explain below why this is necessary.
    \item Even though stateful elements such as LSTM are not used, the network is recurrent because the previous outputs are included as inputs. This is needed to produce output waveforms that look like, for example, a square wave.
\end{itemize}

The layer normalization is defined to be the following:
\begin{equation}
    LN(x_i) = \frac{x_i - mean(x_i, x_{i-1}, \dots, x_{i-T+1})}{\epsilon + std(x_i, x_{i-1}, \dots, x_{i-T+1})},
\end{equation}
where the $mean$ and $std$ are over the time dimension. The subscripts $i, i-1,\dots, x_{i-T}$ are the time indices, with $T$ being the sample size. Subtraction of the mean is necessary because of the large range of values the input might take. For example, the position $p_z$ is fed in mm and can vary from $-170~$mm to $+300~$mm depending on the height of the table, whereas the change in $p_z$ during insertion may only be $20~$mm (i.e., the height of the hole). We also use a large $\epsilon = 0.1$ in order to suppress noise when one of the inputs is mostly static and unchanging over time.



\begin{figure}
  \includegraphics[width=\linewidth]{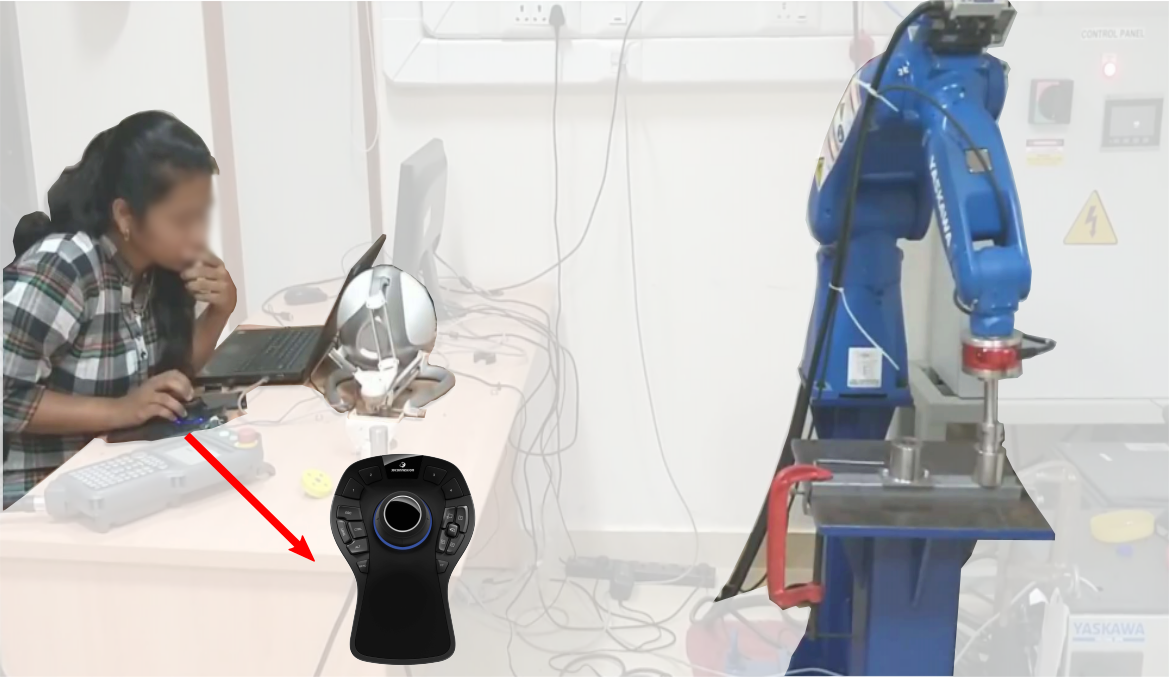}
  \caption{The space mouse used by the human expert to guide the peg into the hole.}
  \label{fig:spacemouse_vj}
\end{figure}

\section{Results}
\label{sec:results}
With the network architecture described, we now discuss the main results. The experiments were performed on a Yaskawa GP8 robot with the MotoFit $6$-axis force sensor. We wrote a program for the YRC1000 controller using the MotoPlus API in C that communicates with a PC via UDP. The average round trip latency between the PC and the YRC1000 controller was less than $1$~ms. The neural network was implemented using Keras on a Linux PC. The sampling rate for collecting the sensor data and to update the target force was $100$~Hz. 
We assume that the position of the hole relative to the base of the robot is known and the peg is already positioned to be just above the hole (with an error of $+/-1~$mm, and with a randomly chosen roll and pitch angle (i.e., between $-2$ and $+2$ degree with respect to the hole). Therefore, a camera is not necessary. However, even if the peg is precisely placed above the hole, plain insertion i.e., applying a downward force on the peg, is not guaranteed to be successful every time. This is due to the fact that the position accuracy of the robot is $>20$~$\mu m$. 
The transimpedance controller serves as the low level controller that receives actuator commands from the policy network. 

\subsection{Data collection}

Collecting expert data proved to be a challenge. One option was to detach the peg from the robot while leaving it connected to the force sensor and to have the expert insert it in the hole. The problem with this approach is that it is difficult to measure what forces the expert is applying on the peg. Instead we chose to display the force sensor readings along with the position of the peg on a screen and to have the expert apply a force on the peg using a space mouse (Fig.~\ref{fig:spacemouse_vj}). The lack of haptic feedback is a major drawback of this approach. Nevertheless, the experts were successful in inserting the peg.

\begin{figure}
  \includegraphics[width=\linewidth]{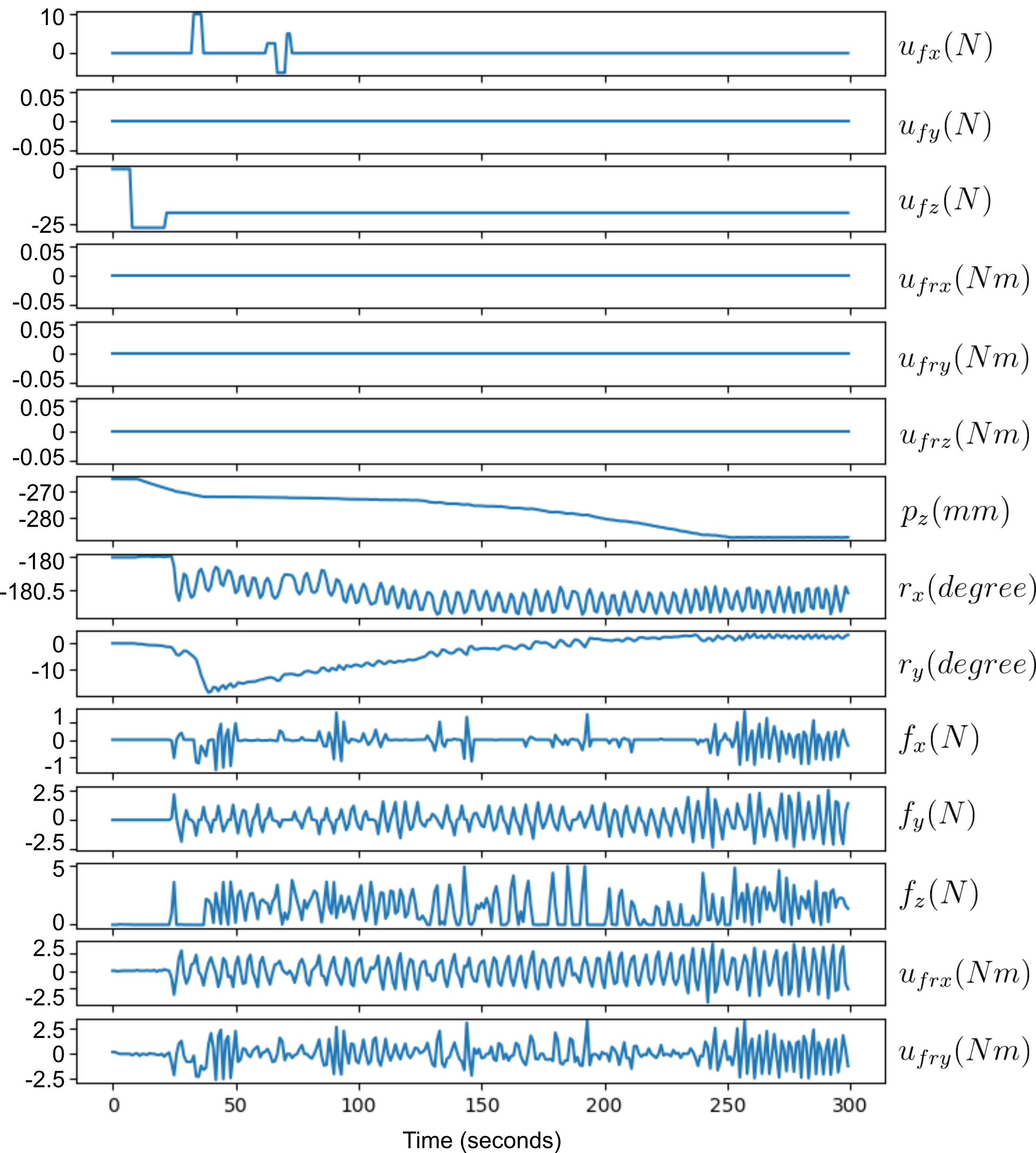}
  \caption{A sample episode where the expert guides the peg into the hole using the space mouse. The plots are described as follows (from top to bottom): force along $x$ direction ($u_{f_x}$), force along $y$ direction ($u_{f_y}$), force along $z$ direction ($u_{f_z}$), moment about $x$ direction ($u_{f_{rx}}$), moment about $y$ direction ($u_{f_{ry}}$), moment about $z$ direction ($u_{f_{rz}}$), vertical position of peg $p_z$, roll of peg $r_x$, pitch of peg $r_y$, force measured along $x$, force measured along $y$, force measured along $z$, moment measured about $x$, and moment measured about $y$. }
  \label{fig:motocap}
\end{figure}

Figure~\ref{fig:spacemouse_vj} shows the expert operating the space mouse to guide the peg into the hole. A constant downward force is applied during insertion. The expert can apply forces in the $X$ and $Y$ direction using the space mouse. Although a torque in $R_z$ can be applied, none of the experts did so. A sample episode where the expert successfully inserts the peg is shown in Fig.~\ref{fig:motocap}. We collected 8 such episodes for training.

\subsection{GAIL Training}

\begin{figure*}
	\centering
	\includegraphics[height=0.18\textwidth]{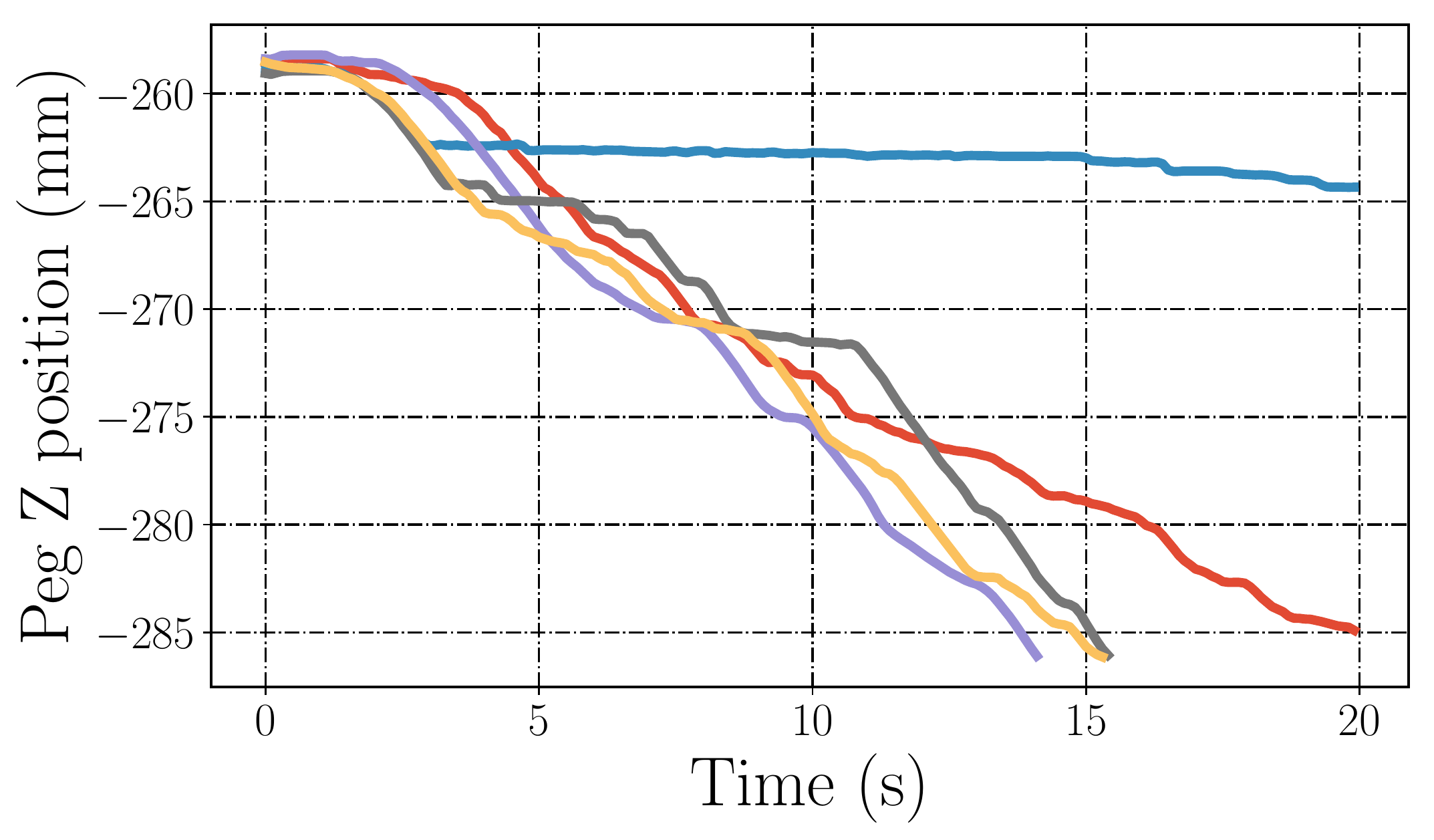}
	\includegraphics[height=0.18\textwidth]{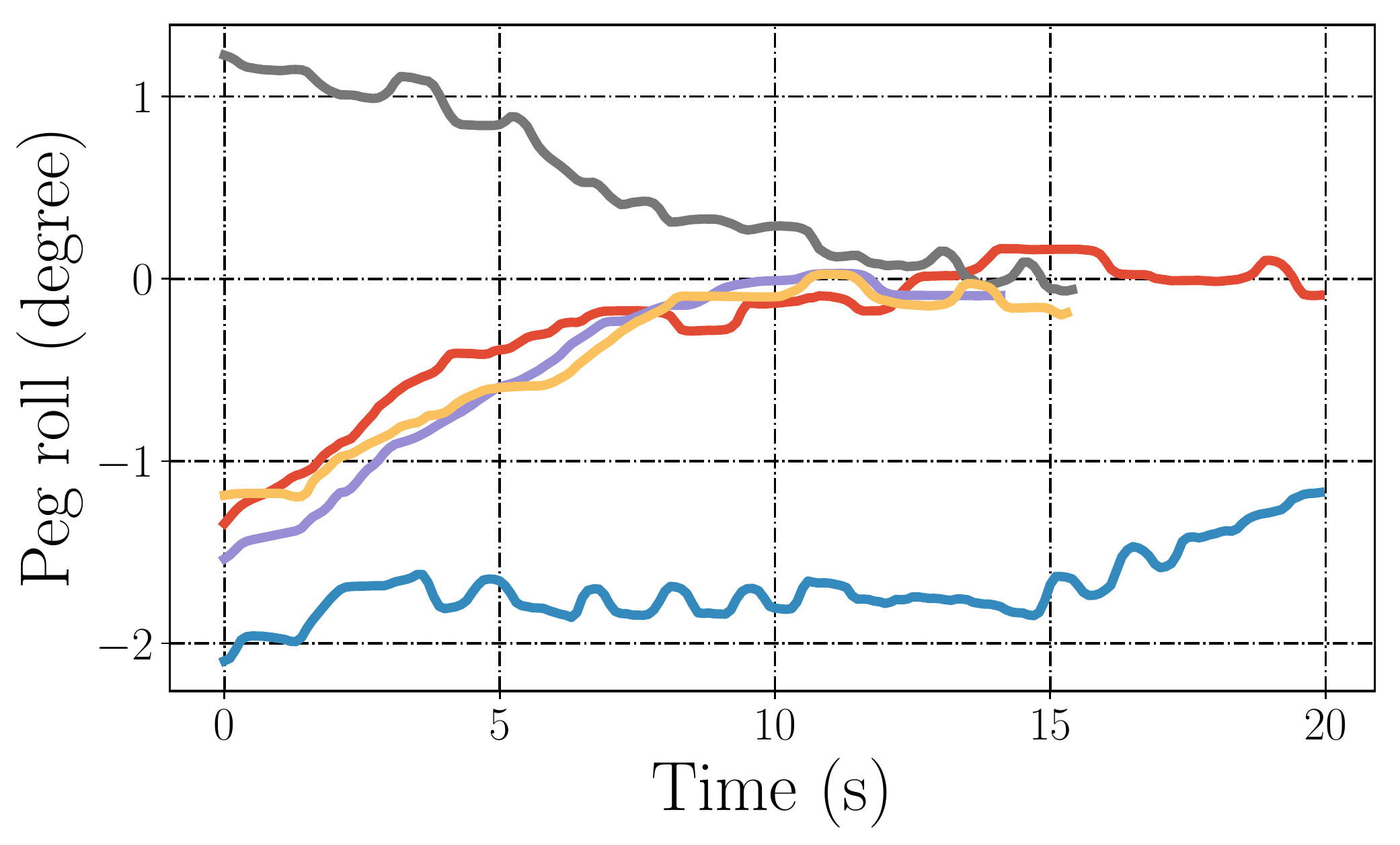}
	\includegraphics[height=0.18\textwidth]{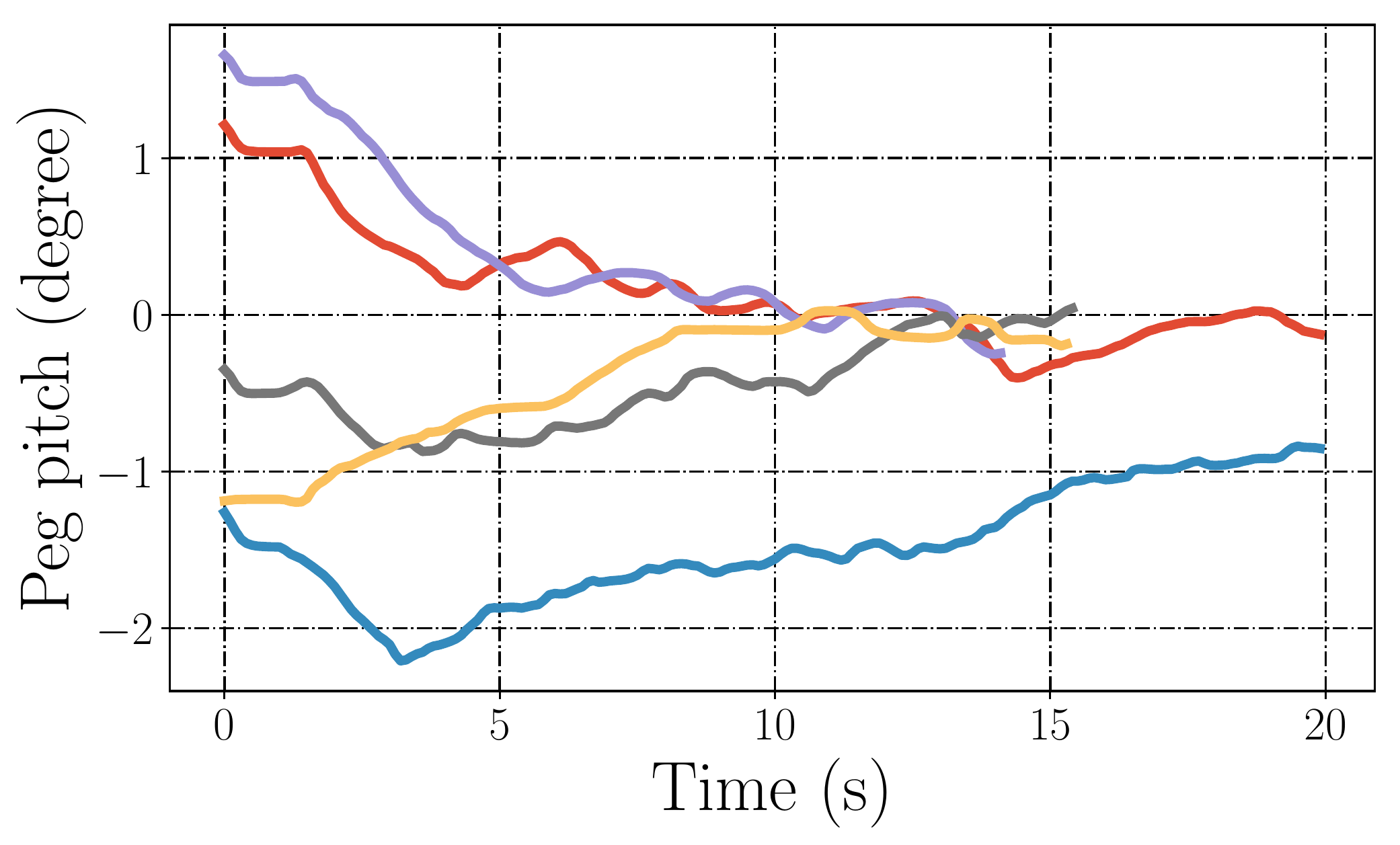}
	\caption{Position of the peg $p_z$, roll of the peg $p_{r_x}$, and pitch angle of the peg $p_{r_y}$ are given here for $5$ episodes (randomly chosen from the $20$). It can be verified that the initial pose varies from $-2$ to $2$ degrees.}
	\label{fig:pz}
\end{figure*}

Roughly $500$ samples from the MotoFit algorithm were used for Generator. Discriminators are trained with cross-entropy loss, and the generators are trained with PPO loss \cite{schulman2017proximal}. After every episode equal number of samples are chosen both from the generated and expert data, and the discriminator is trained for about $100-500$ iterations. Similarly, after every episode, the discriminator is evaluated for the samples and used as reward for the generator. With these rewards, the generator is trained via PPO. It is worth noting that the frequency of update for the generator is much higher than that of the discriminator. This is due to the fact that the discriminator tends to learn faster than the generator. The above training framework is run for about $20$ episodes in the robot, and the results are obtained. Fig. \ref{fig:pz} shows the trajectories of vertical position, roll and pitch angles of the peg during some of the training episodes. It can be verified that the ``wiggle" mode with a downward force is turned on whenever the peg is stuck.

\begin{figure*}
	\centering
	\includegraphics[height=0.18\textwidth]{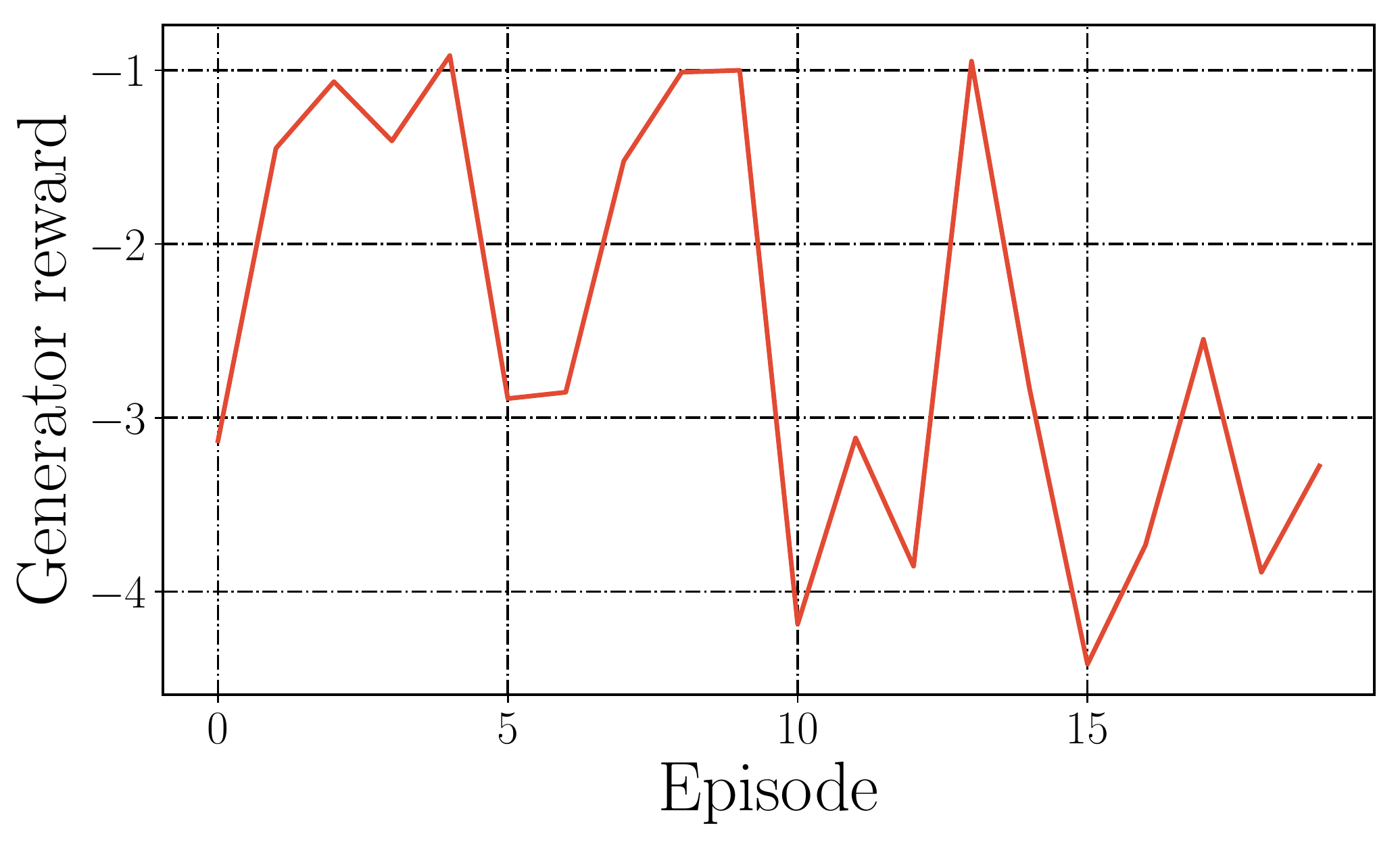}
	\includegraphics[height=0.18\textwidth]{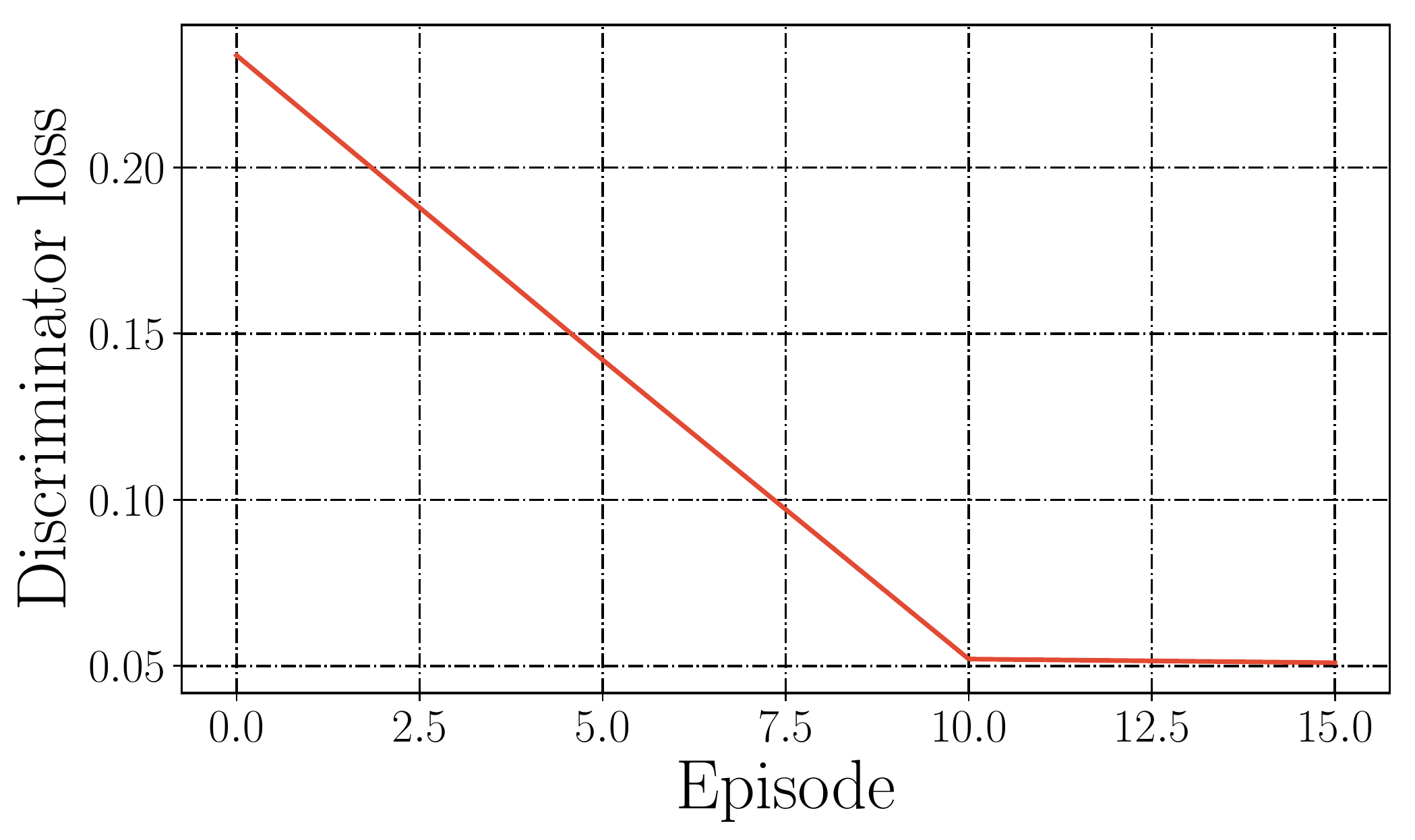}
	\includegraphics[height=0.18\textwidth]{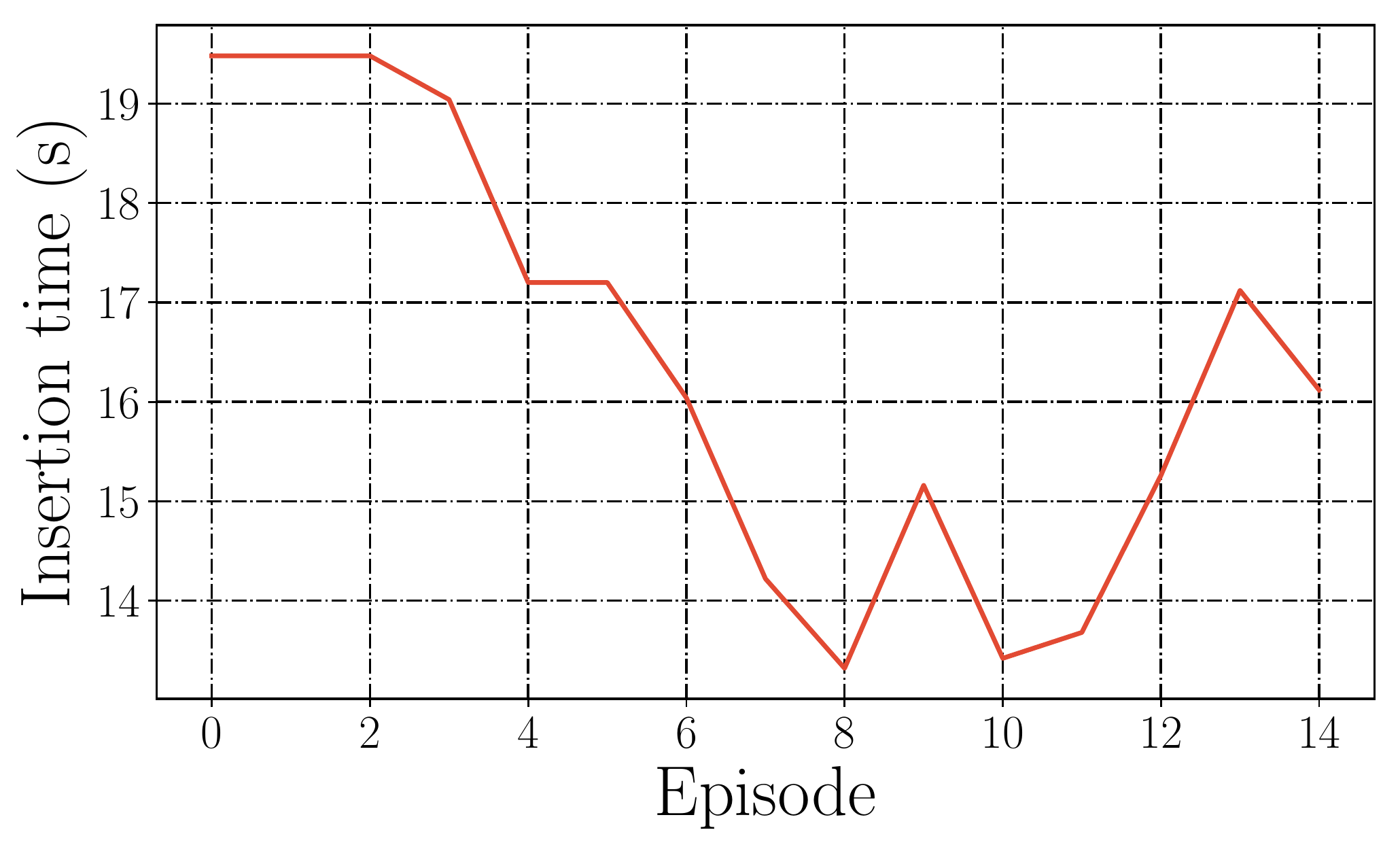}
	\caption{Figures showing the generator reward and the discriminator loss and the mean insertion time (averaged over $5$ consecutive episodes) for twenty episodes. The generator reward oscillates, while the discriminator loss stabilizes after approximately ten episodes. More importantly, the mean insertion time is decreasing over time.}
	\label{fig:gd}
\end{figure*}

\begin{figure*}
	\centering
	\includegraphics[height=0.18\textwidth]{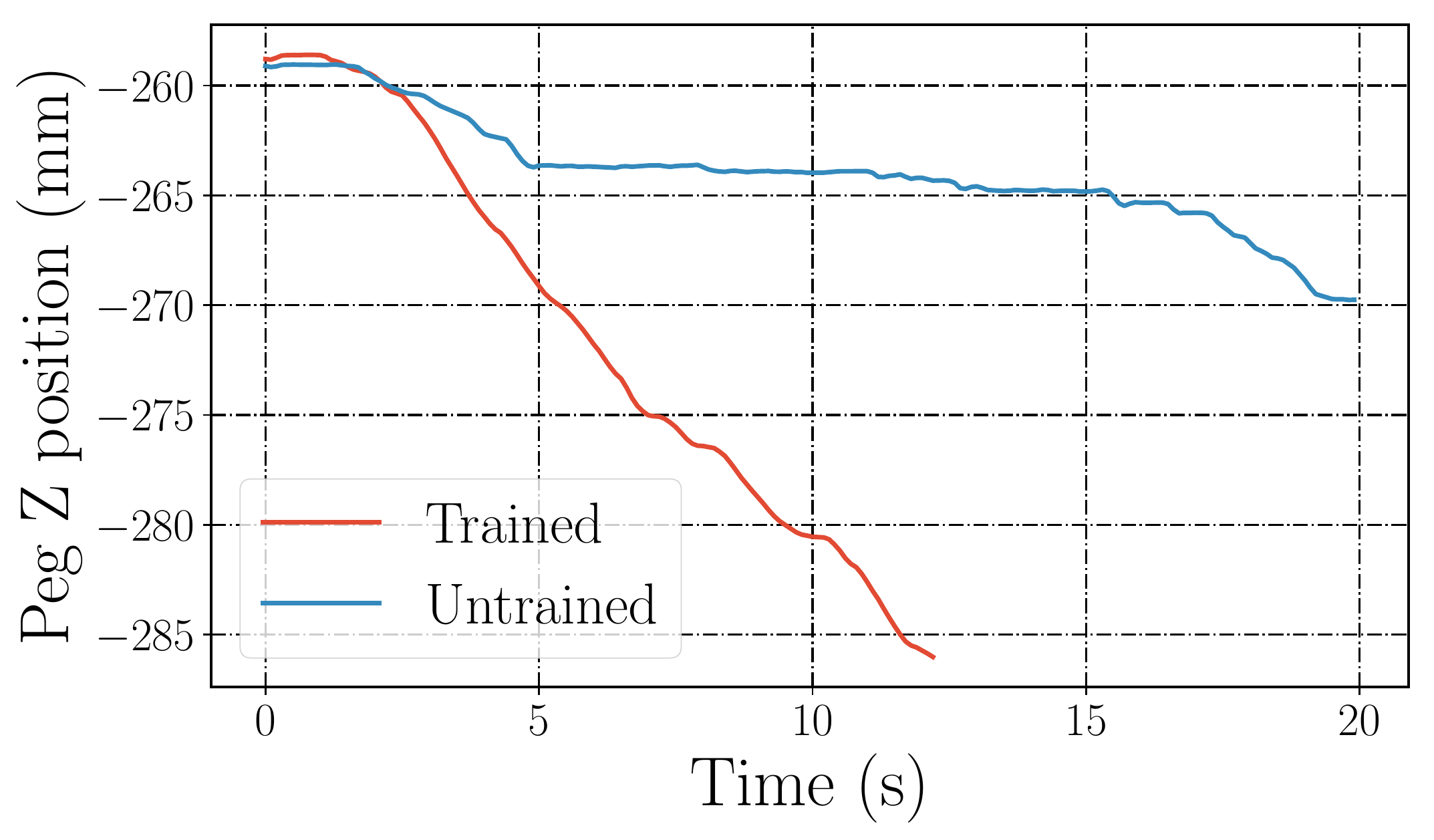}
	  \includegraphics[height=0.18\textwidth]{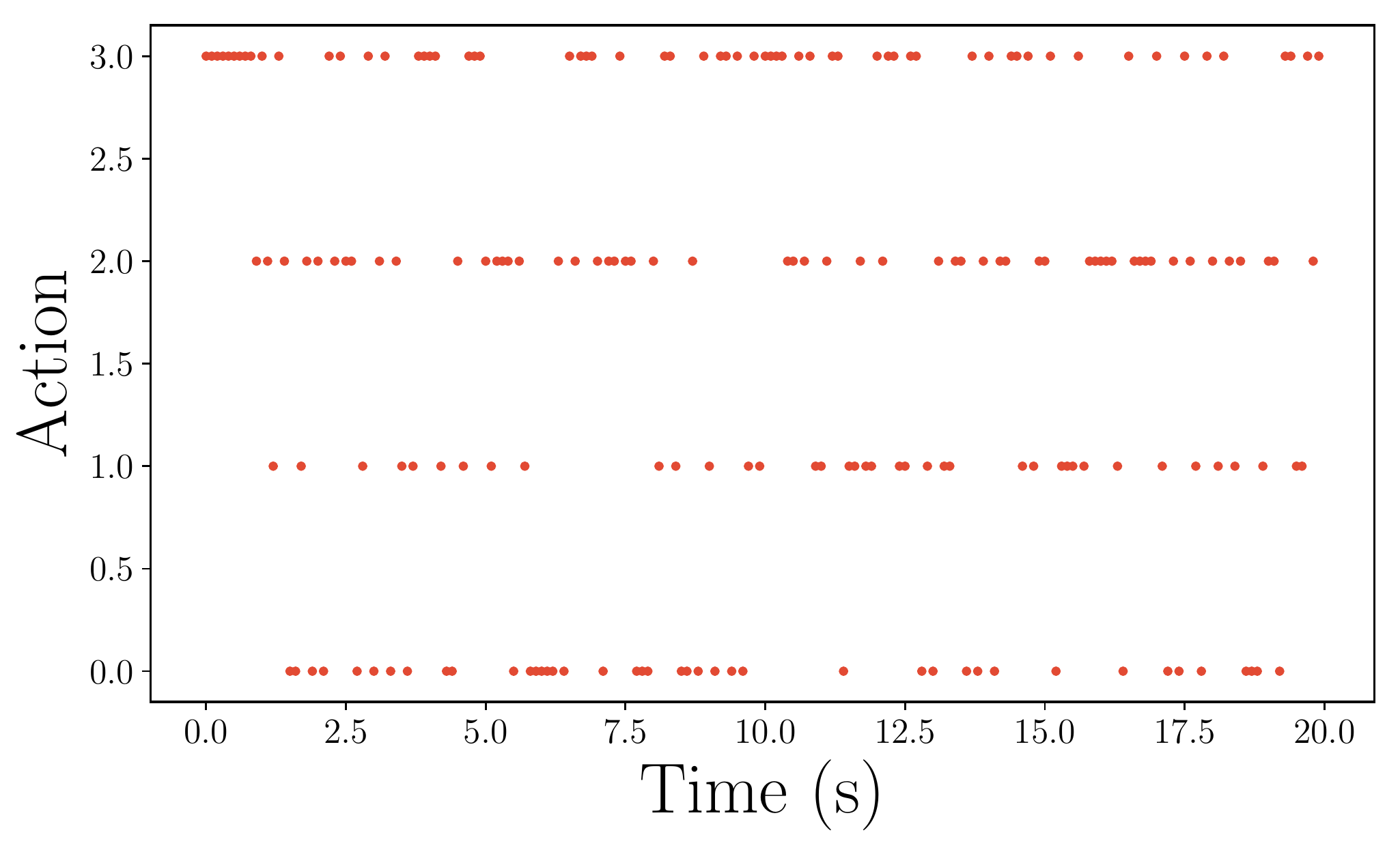}
  \includegraphics[height=0.18\textwidth]{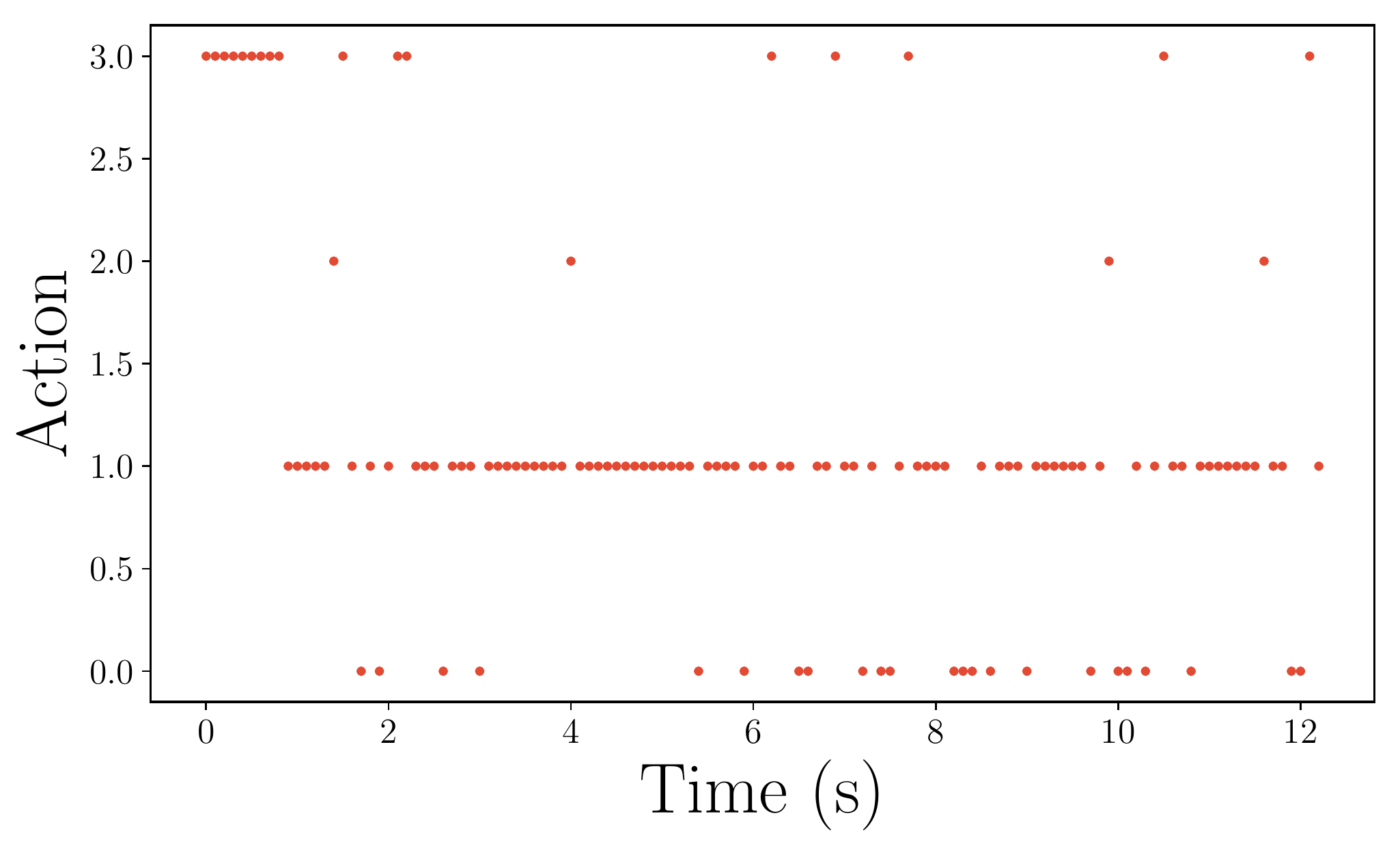}
	\caption{Left figure is showing the comparison between the results with the untrained and trained networks. Actions from an untrained (middle) and a trained (right) network are also shown. The actions from the untrained network are arbitrary, while the trained network yields a more meaningful policy.
		}
	\label{fig:trainuntraincomp}
\end{figure*}

Figs. \ref{fig:gd} show the generator reward and the discriminator loss as a function of the episodes. It is worth noting that the generator and discriminator are playing  a game, and there is no termination condition. Both the losses must stabilize to a constant value (or oscillates), which are nonzero. Fig. \ref{fig:trainuntraincomp} shows the comparison between the untrained and trained networks. As shown by Fig. \ref{fig:trainuntraincomp}, the action values applied before training are arbitrary, and those applied after training are more meaningful. 
It can be verified that the insertion time is also improved (from Fig. \ref{fig:gd}) after about eight episodes. A video demonstrating the training of peg-in-hole is provided in this link: \url{https://youtu.be/V7WQ9MQK17s}.

\section{Conclusion}
\label{sec:conc}
We showed that high precision assembly tasks such as peg-in-hole insertions with small clearances can be trained using imitation learning. Imitation learning is more sample efficient than reinforcement learning and does not require reward shaping. We used generative methods for imitating the expert policy. 
However, it is worth noting that gathering expert data for such tasks still remains a challenge. Future work will involve using better data collection techniques and performing imitation learning on a wider array of tasks.

\section*{ACKNOWLEDGMENT}
We would like to thank Prof. Shalabh Bhatnagar for helpful discussions, and Vaijayanti Ballolli, Raviteja Upadrashta, and Mouleeshwara Reddy for their assisting with the Yaskawa GP8 robot.

\bibliographystyle{IEEEtran}
\bibliography{sample-base}

\end{document}